\documentclass[conference]{IEEEtran}
\IEEEoverridecommandlockouts
\usepackage{cite}
\usepackage{amsmath,amssymb,amsfonts}
\usepackage{algorithmic}
\usepackage{graphicx}
\usepackage{textcomp}
\usepackage{xcolor}
\def\BibTeX{{\rm B\kern-.05em{\sc i\kern-.025em b}\kern-.08em
    T\kern-.1667em\lower.7ex\hbox{E}\kern-.125emX}}
\begin{document}

\title{STF: Spatial Temporal Fusion for Trajectory Prediction
}


\author {
    Pengqian Han,\textsuperscript{\rm 1}
    Jiamou Liu, \textsuperscript{\rm 1}
    Tianzhe Bao, \textsuperscript{\rm 2}
    Yifei Wang, \textsuperscript{\rm 1}\\

    \textsuperscript{\rm 1} The University of Auckland, Auckland, New Zealand\\
    \textsuperscript{\rm 2} University of Health and Rehabilitation Sciences, Qingdao, China\\
}

\maketitle

\begin{abstract}

Trajectory prediction is a challenging task that aims to predict the future trajectory of vehicles or pedestrians over a short time horizon based on their historical positions. The main reason is that the trajectory is a kind of complex data, including spatial and temporal information, which is crucial for accurate prediction. Intuitively, the more information the model can capture, the more precise the future trajectory can be predicted. However, previous works based on deep learning methods processed spatial and temporal information separately, leading to inadequate spatial information capture, which means they failed to capture the complete spatial information. Therefore, it is of significance to capture information more fully and effectively on vehicle interactions. In this study, we introduced an integrated 3D graph that incorporates both spatial and temporal edges. Based on this, we proposed the integrated 3D graph, which considers the cross-time interaction information. In specific, we design a Spatial-Temporal Fusion (STF) model including Multi-layer perceptions (MLP) and Graph Attention (GAT) to capture the spatial and temporal information historical trajectories simultaneously on the 3D graph. Our experiment on the  ApolloScape Trajectory Datasets shows that the proposed STF outperforms several baseline methods, especially on the long-time-horizon trajectory prediction. Our code is available at https://github.com/pengqianhan/STF-Spatial-Temporal-Fusion-for-Trajectory-Prediction

\end{abstract}

\begin{IEEEkeywords}
Graph Neural Network, Trajectory Prediction, Spatial-Temporal Data Mining
\end{IEEEkeywords}

\section{Introduction}
The trajectory prediction is to accurately predict the future path of an object based on its historical position. Trajectory data is a type of spatio-temporal data that contains both spatial and temporal information. Spatial information pertains to the interaction between objects due to their relative position, movement tendency, or relative velocity. Besides physical interaction, there is social interaction between vehicles. Because traffic can not consider all the driving behavior and the drivers do not always follow the traffic rules, sometimes they prefer to reach their destination in the shortest time\cite{wang2022social-social_interaction_survey}. Temporal information refers to the time at which the object is at a specific location or moving from one location to another.

\begin{figure}[htbp]
    \centering
    
    \begin{minipage}[b]{0.45\linewidth}
        \centering
        \includegraphics[width=\linewidth]{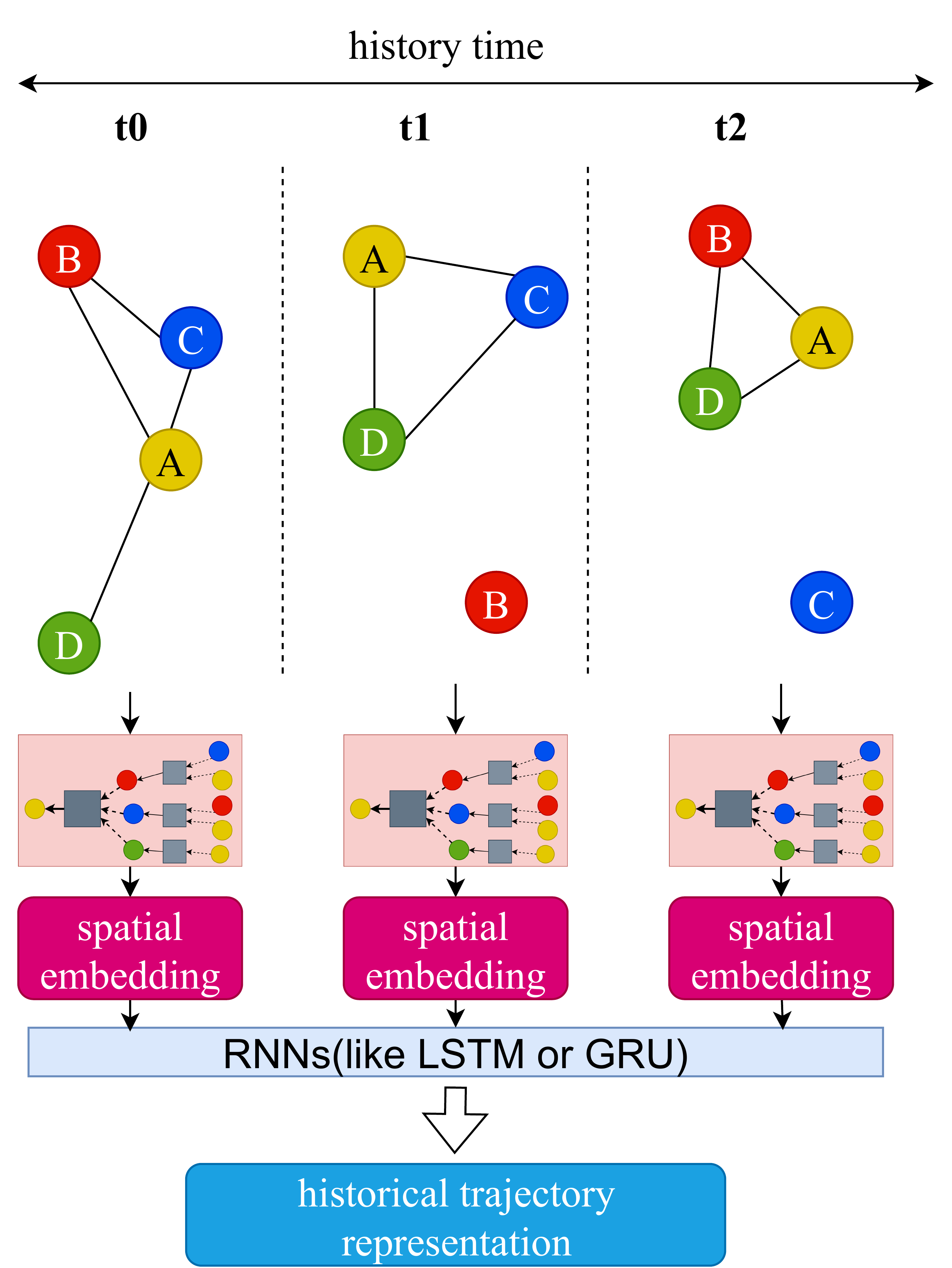}
        \caption{2D graph in previous works}
        \label{fig:limitation}
    \end{minipage}
    \hfill
    \begin{minipage}[b]{0.45\linewidth}
        \centering
        \includegraphics[width=\linewidth]{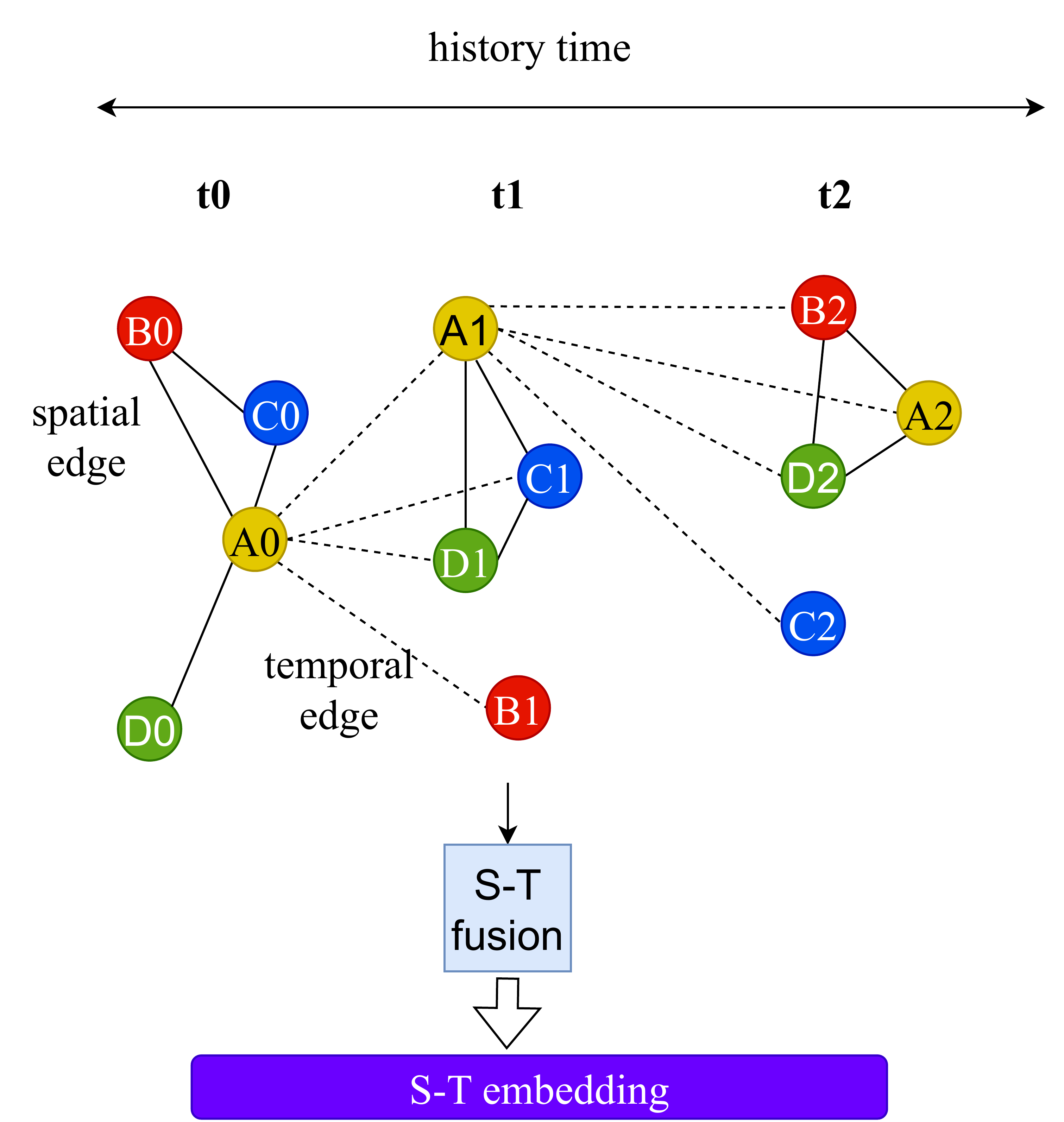}
        \caption{Integrated 3D graph in S-T Fusion}
        \label{fig:3dgraphfusion}
    \end{minipage}
    
\end{figure}

The prediction of the trajectories of surrounding vehicles and pedestrians is vital for an autonomous vehicle, enabling it to make informed decisions to avoid collisions. On-board and off-board sensors are used to detect other vehicles, lanes, and crosswalks and generate large amounts of data that can be used to train high-performance deep learning models. These models can predict feasible and efficient trajectories, improving safety and traffic conditions.
Machine learning methods such as Hidden Markov Models (HMMs), Support Vector Machines (SVMs), and Dynamic Bayesian Networks (DBNs) have been used to address this problem. However, these approaches are limited to short time horizons, like 2 seconds, and their prediction performance is unsatisfactory. Deep learning is a subfield of machine learning that utilizes multiple layers in its neural networks and has demonstrated significant efficacy with the increase in data volume\cite{zhang2021dive-d2l}. It enables computers to recognize objects from images or translate languages. When applied to trajectory prediction, deep learning methods have proven to outperform conventional machine learning methods. In addition, some work based on deep learning methods like \cite{gupta2018social-socialgan, alahiSocialLSTMHuman2016a-sociallstm, mohamed2020social-social-stgcnn, yu2020spatio-STAR, Mendieta_Tabkhi_2022-CARPe} have demonstrated exceptional performance in long-term trajectory prediction tasks. However, these works process the spatial and temporal separately. As depicted in Fig. \ref{fig:limitation}, certain studies, such as CS LSTM\cite{deoConvolutionalSocialPooling2018a-cslstm}, only consider spatial information at the final time step. On the other hand, GRIP++\cite{li2019grip++} employs graph neural networks (GNNs) to capture spatial information and subsequently utilize customized convolutional neural networks (CNNs) to capture temporal information. Additionally, Social LSTM \cite{alahiSocialLSTMHuman2016a-sociallstm} and Starnet \cite{zhu2019starnet} adopt pooling methods to capture the spatial characteristics of historical trajectories before employing LSTM \cite{hochreiter1997long-lstm} for capturing temporal information. All of these methods missed the interaction information across the time frames during history time. The vehicles on the road interact with each other, occurring within the same time step and across multiple time steps. Previous research has focused on the interaction between agents. However, they have often treated spatial and temporal information in isolation, resulting in a lack of consideration for interaction related to agents at different time intervals. In this work, we define an integrated 3D graph with spatial edges between the nodes at the same time stamp and temporal edges across the time stamp illustrated in Fig. \ref{fig:3dgraphfusion}. After defining the 3D graph, all kinds of vehicle interactions are considered, enhancing the model to capture more information from the historical trajectory. Finally, we propose a \textbf{S}patial-\textbf{T}emporal \textbf{F}usion (STF) model to predict the future trajectory. Our approach effectively captures more historical trajectory data information than previous models. We conducted several experiments that demonstrated significantly improved performance over existing methods. The code is available on https://github.com/. The main contribution of this work can be summarized as follows:
\begin{itemize}
    \item This study introduces an integrated 3D graph that incorporates both spatial and temporal edges. The spatial edges represent the spatial interaction between vehicles within the same time frames, while the temporal edges indicate cross-time frame interactions, referred to as spatial-temporal interactions.
    
    \item A high-performance trajectory prediction model called STF is proposed in this work. This model effectively captures a wider range of information from historical trajectories, including not only spatial interactions but also spatial-temporal interactions. Additionally, it successfully incorporates temporal information into its predictions.
    
    \item Several experiments are conducted to evaluate the performance of the proposed approach, demonstrating superior results compared to previous research papers.
\end{itemize}

\section{Related works}
The methods in previous works are very similar for capturing the temporal information, most of them choosing the RNNs-based models, including LSTM and GRU. However, they apply various algorithms to capture the  spatial information of trajectory. These deep learning methods for trajectory prediction can be classified into three categories.

\subsection{Pooling and CNNs based method}
To capture information regarding pedestrian interaction, Social LSTM \cite{alahiSocialLSTMHuman2016a-sociallstm} implemented a grid-based social pooling technique between LSTM cells. Although social pooling addresses the issue of varied numbers of neighboring pedestrians, it does not consider different weights assigned to neighbor agents. TraPHic \cite{ma2019trafficpredict-apollodataset} also utilizes social pooling to capture interaction information. The operation regions in TraPHic's social pooling consist of a Horizon map and a Neighbor Map, which enable TraPHic to assign different weights to different agents via social pooling in two maps. It improves the Social LSTM but still misses the cross-time interaction. In CS LSTM \cite{deoConvolutionalSocialPooling2018a-cslstm}, shared weights LSTM encoder output is used to make a social tensor, which is put through Convolutional Social Pooling. The Convolutional Social Pooling captures interaction at the last time of the encoder, but it ignores long-range interactions. Social GAN \cite{gupta2018social-socialgan} designs a Pooling Module to capture the interaction of all pedestrians in a scene, whereas Social Pooling \cite{alahiSocialLSTMHuman2016a-sociallstm} can only handle people in an $m \times n$ grid. The relative positions between the target pedestrian and others are fed into an MLP whose output is concatenated with the target's hidden states. The resulting tensor is then processed by another MLP following a max pooling layer.

\subsection{GNNs based method}
The Social-STGCNN model \cite{mohamed2020social-social-stgcnn} utilizes Graph Convolutional Networks (GCNs) to capture spatial information and model pedestrian interactions based on a graph defined at each historical frame. Similarly, GRIP++ \cite{li2019grip++} defines graphs that are then processed by GCNs and 2D temporal convolution layers to capture spatial and temporal information from observed tracks. GSTCN \cite{sheng2022graph-gstcn} defines a complete graph with weighted edges based on reciprocal distances between nodes to model vehicle interactions. ReCoG \cite{mo2020recog} uses a directed graph where edges connect vehicles and the target if their distance is below 20 meters. RSBG \cite{sun2020recursive-rsbg} trains a Recursive Social Behavior Generator to output weighted edges representing latent social relationships, which are then fed into GCN layers. SGCN \cite{shi2021sgcn} addresses the issue of superfluous edges by first defining a dense undirected graph and then converting it into a sparse directed graph to model pedestrian interactions. The Social-BiGAT model \cite{kosaraju2019social-social-bigat} employs GAT and a latent encoder to process hidden states and generate latent noise, enhancing model multimodality. All the graph-based deep learning methods miss the traffic participants' interaction cross time because they define the graph at every frame, not the whole historical time.
\subsection{Attention based method}
LaneGCN \cite{liang2020learning-lanegcn} utilizes attention mechanisms to learn interactions among lanes, actors, and actors themselves. VectorNet \cite{gao2020vectornet-Vectornet} employs a hierarchical graph, with subgraphs representing agent trajectories and road features, which are processed using an attention mechanism. HiVT \cite{zhou2022hivt} adopts a hierarchical approach by capturing local interactions based on relative positions using an attention mechanism. It then applies a temporal transformer to generate spatial and temporal features of the central agent. Another attention mechanism encodes lane information into keys and values using an MLP and calculates pairwise global interaction. These papers treated temporal information and social interaction separately. The ignorance of cross-time interaction leads to reduced model performance. 
\section{PROBLEM FORMULATION}\label{problem formulation}
Let $N$ denote the number of traffic agents, $T_{his}$ represent the history time, and $T_{pred}$ indicate the prediction time. The input history trajectories set of agents is denoted by $\mathbf{X}=[\mathbf{x}^{1},\mathbf{x}^{2},...,\mathbf{x}^{T_{his}}]$, where $\mathbf{x}^{t}$ represents the coordinates of N agents at time $t$, i.e., $\mathbf{x}^{t}=[x_0^t,y_0^t,x_1^t,y_1^t,...,x_{N-1}^t,y_{N-1}^t]$.

The prediction trajectories are defined as $\mathbf{Y}$, which includes future positions from time $( T_{his} + 1 )$ to $( T_{his} + T_{pred})$, i.e., $\mathbf{Y}=[\mathbf{y}^{T_{his}+1},\mathbf{y}^{T_{his}+2},...,\mathbf{y}^{T_{his}+T_{pred}}]$.

\section{Spatial temporal fusion model }
\subsection{Scheme of model}
\begin{figure}
    \centering
    \includegraphics[width=\linewidth]{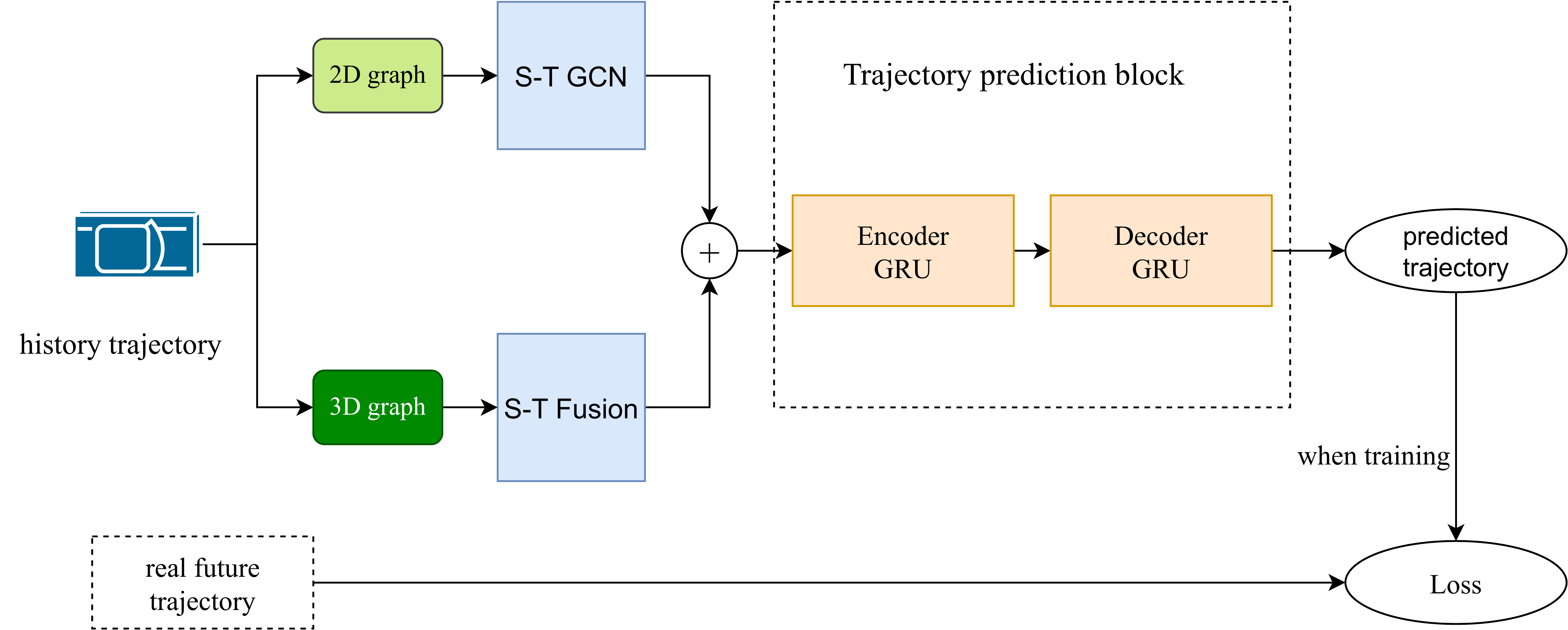}
    \caption{Scheme of ST fusion model. The 2D graph is defined on all historical time stamps, while the 3D graph is defined based on the entire historical time stamps, including spatial edges and temporal edges}
    \label{fig:Architecture of model}
\end{figure}

Fig. \ref{fig:Architecture of model} illustrates the architecture of the proposed model, which utilizes the  historical trajectory of traffic agents, including cars, bikes, and pedestrians, as input. The S-T fusion block extracts both spatial and temporal information simultaneously from this input. Meanwhile, the ST-GCN block separately captures spatial and temporal information. These two blocks' outputs are concatenated before being fed into the Seq2Seq network \cite{sutskever2014sequence-seq2seq} for predicting future trajectories within a time period of $T_{pred}$.

\subsection{Integrated 3D graph definition}

In order to process the spatial and temporal information simultaneously, this work creates a big graph where each node represents a traffic agent, such as bikes, cars, and pedestrians. An edge is created between two nodes at each time stamp if their distance is shorter than the threshold $D_{close}$. These edges are considered \textbf{spatial edges}. Additionally, edges between nodes at adjacent time stamps are defined as \textbf{temporal edges} used to capture interactions across different time stamps. Fig. \ref{fig:3dgraphfusion} displays the spatial and temporal edges. Although other nodes' temporal edges are comparable, they are not depicted in the illustration. Once the historical trajectories are compiled into an integrated 3D graph, an adjacency matrix can be generated, shown in Table \ref{tab:adjacent matrix of biggraph}.

\begin{table}[htbp]
            \centering
            \caption{Adjacent matrix of biggraph}
            \label{tab:adjacent matrix of biggraph}
            \resizebox{0.8\linewidth}{!}{
            \begin{tabular}{c|cccc|cccc|cccc}
                & A0 & B0 & C0& D0 &A1 & B1 & C1 & D1 & A2 & B2 & C2 & D2 \\
            \hline
            A0 & 0 & 1 & 1 &  1 & 1  & 1  & 1  & 1 & 0  & 0  & 0  & 0 \\
            B0 & 1 & 0 & 1 &  0 & 1  & 1  & 1  & 1 & 0  & 0  & 0  & 0\\
            C0 & 1 & 1 & 0 &  0 & 1  & 1  & 1  & 1 & 0  & 0  & 0  & 0\\
            D0 & 1 & 0 & 0 &  0 & 1  & 1  & 1  & 1 & 0  & 0  & 0  & 0\\
            \hline
            A1 & 1 & 1 & 1 &  1 & 0  & 0  & 1  & 1 & 1  & 1  & 1  & 1\\
            B1 & 1 & 1 & 1 &  1 & 0  & 0  & 0  & 0 & 1  & 1  & 1  & 1\\ 
            C1 & 1 & 1 & 1 &  1 & 1  & 0  & 0  & 1 & 1  & 1  & 1  & 1\\
            D1 & 1 & 1 & 1 &  1 & 1  & 0  & 1  & 0 & 1  & 1  & 1  & 1\\
            \hline
            A2 & 0 & 0 & 0 &  0 & 1  & 1  & 1  & 1 & 0  & 1  & 0  & 1\\
            B2 & 0 & 0 & 0 &  0 & 1  & 1  & 1  & 1 & 1  & 0  & 0  & 1\\ 
            C2 & 0 & 0 & 0 &  0 & 1  & 1  & 1  & 1 & 0  & 0  & 0  & 0\\
            D2 & 0 & 0 & 0 &  0 & 1  & 1  & 1  & 1 & 1  & 1  & 0  & 0 \\
            \end{tabular}}

        \end{table}

\subsection{Spatial-Temporal Fusion block}
The model takes the location $(x_i^t,y_i^t)$ of traffic agents during a historical period $T_{his}$ as input.  To better represent this input, two layers of MLP are utilized to increase its dimension. This results in an output dimension of 16 for the MLP, as demonstrated in Equation (\ref{eq:mlp}).

\begin{equation}
    e_i^t=MLP(x_i^t,y_i^t),where \ e_i^t \in \mathbb{R}^F,F=16
    \label{eq:mlp}
\end{equation}
The output of the MLP is provided as input of  two layers of GAT, allowing for simultaneous capture of both spatial and temporal information. The input of graph attention layers is Equation (\ref{eq:gat_input})
\begin{equation}
    h=\left\{e_1^1,e_2^1,...,e_N^1,e_1^2,e_2^2,...,e_N^2,....,e_1^{T_{his}},e_2^{T_{his}},...,e_N^{T_{his}}\right\}
    \label{eq:gat_input}
\end{equation}
where $e_n^t$ is the embedding of $nth$ node at time $t$.

\begin{figure}
    \centering
    \includegraphics[width=0.6\linewidth]{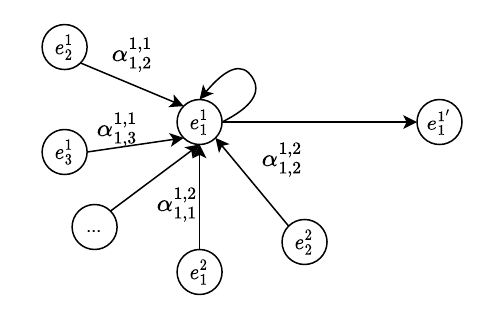}
    \caption{An illustration of graph attention layer. It allows a node
to assign different importance to different nodes within the threshold and in the adjacent time stamps. The new feature of nodes is aggregated from them. }
    \label{fig:gat_biggraph}
\end{figure}

{\footnotesize
\begin{equation}
\alpha_{u,v}^{i,j} = \frac{\exp\left(\text{LeakyRelu}\left( a^T\left[We_u^i \bigoplus We_v^j\right]\right)\right)}{\sum_{k \in \mathbb{N}(u)} \sum_{l \in \left\{i-1,i,i+1\right\}} \exp\left(\text{LeakyRelu}\left( a^T\left[We_u^i \bigoplus We_k^l\right]\right)\right) } 
\label{eq:gat}
\end{equation}
}
where $\alpha_{u,v}^{i,j}$ is the attention weights between node $u$ at time $i$ and node $v$ at time $j$. $W \in \mathbb{R}^{F'\times F}$ is a shared trainable matrix for linear mapping. $F'$ is the output feature dimension, and $F$ is the input feature dimension. ${\bigoplus}$ denotes the concatenation operation. $\mathbb{N}(u)$ means the neighbor nodes of $u$ within a threshold at the same time. $a \in \mathbb{R}^{2F'}$ is a weight vector.

The updated representation of node $u$ at time $i$, denoted as $e_u^{i'}$, is obtained by computing the attention weights of adjacent nodes on target nodes, as specified in Equation (\ref{eq:gat_out}).

\begin{equation}
    e_u^{i'}=\sum _{k \in \mathbb{N}(u)} \sum _{l \in \left\{i-1,i,i+1\right\}}\alpha_{u,v}^{i,j}We_u^i
    \label{eq:gat_out}
\end{equation}

\begin{figure}
    \centering
    \includegraphics[width=\linewidth]{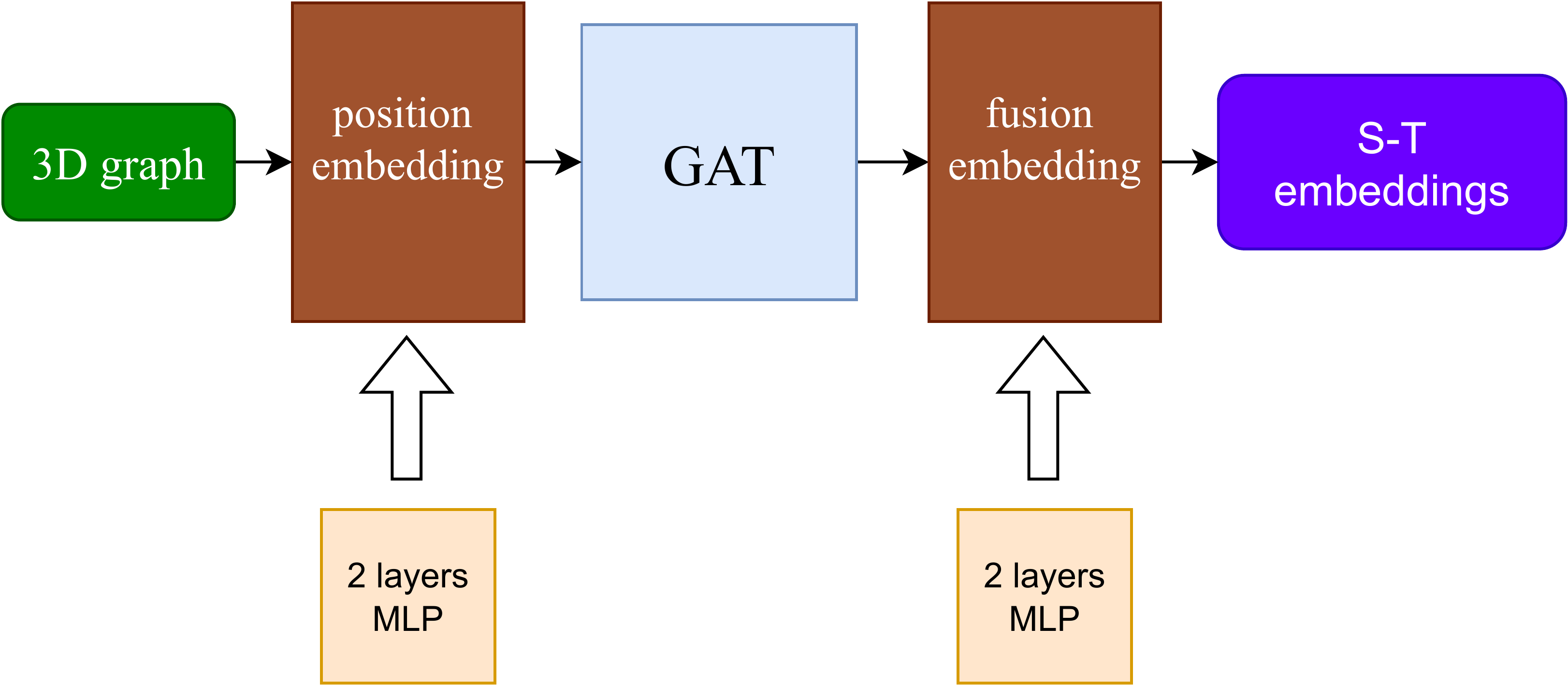}
    \caption{Spatial-Temporal fusion block}
    \label{fig:st-fusion block}
\end{figure}

\subsection{Spatial-Tempotal GCN block}
\begin{figure}
    \centering
    \includegraphics[width=\linewidth]{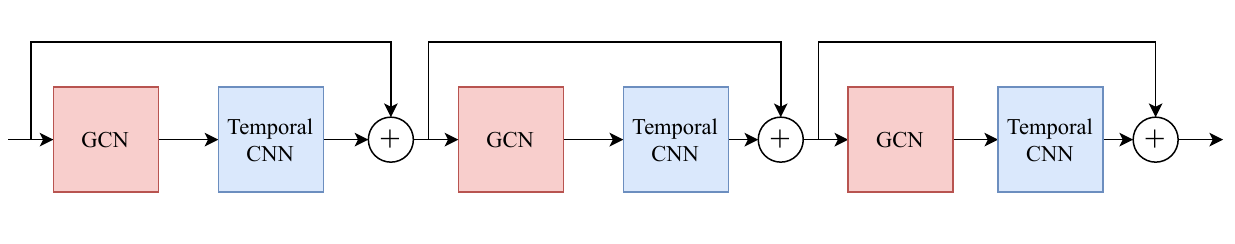}
    \caption{Spatial-Tempotal GCN block}
    \label{fig:stgcn}
\end{figure}

The block records the historical $x,y$ coordinates of $N$ traffic agents and processes spatial and temporal information separately. To capture interactions, input data is first fed into a GCN layer, followed by a temporal CNN layer to capture time information. The Spatial-Temporal GCN block consists of three ST-GCN layers, as illustrated in Fig. (\ref{fig:stgcn}).
\subsection{Trjectory prediction block}
The trajectory prediction block is a conventional sequence-to-sequence (seq2seq) model comprising GRUs as its components. The seq2seq model excels at processing sequence data and has been widely utilized in prior studies such as \cite{alahiSocialLSTMHuman2016a-sociallstm} and \cite{song2020pip}. Previous experiments have demonstrated that the seq2seq model exhibits high performance in trajectory prediction tasks. Furthermore, trajectory prediction belongs to the domain of autoregressive tasks, which are well-suited for LSTMs and GRUs. In this study, we employ the seq2seq model to forecast future trajectories, building upon existing research.
\section{Experiments}
We trained our model on a Linux server running Ubuntu 22.04.1 LTS with AMD Ryzen Threadripper PRO 3995WX 64-Cores, 512GB RAM, and one NVIDIA GeForce RTX 3090. 
\subsection{Dataset}
We train and test our model on the ApolloScape Trajectory Dataset \cite{ma2019trafficpredict-apollodataset}, whose framerate are two frames per second. The training sequence lasts for 53 minutes, while the test sequence lasts for 50 minutes. The dataset contains two files, one for training and one for testing. In the training zip file, each file is a 1-minute sequence, and each line in the file contains frame identification, object identification, object type, position x, position y, position z, object length, object width, object height, and heading. The traffic in the dataset is classified into five categories: small vehicles, large vehicles, pedestrians, motorcyclists, bicyclists, and others. The location is based on global coordinates. For the Apollo challenge, the observation time was set to 2 seconds. However, some studies \cite{li2019grip++}\cite{li2019grip} have extended the observation time to 3 seconds, along with a prediction time of 3 seconds. The testing files have the same data structure as the training files, which are used to evaluate the performance of the model.


\subsection{Metrics}
\begin{table}[h]
\small
\centering
\caption{Competition Results On ApolloScape Trajectory Dataset.}
\label{tab:ade-fde}
\resizebox{\linewidth}{!}{

\begin{tabular}{|c|c|c|c|c|c|c|c|c|}
\hline
Method& WSADE & $\text{ADE}_\text{v}$ & $\text{ADE}_\text{p}$  & $\text{ADE}_\text{b}$  & WSFDE  & $\text{FDE}_\text{v}$ & $\text{FDE}_\text{p}$ & $\text{FDE}_\text{b}$  \\ \hline
TafficPredict \cite{ma2019trafficpredict-apollodataset} & 8.5881         & 7.9467         & 7.1811         & 12.8805        & 24.2262        & 12.7757        & 11.1210        & 22.7912        \\ 
GRIP++ \cite{li2019grip++}        & 2.514          & 3.948          & 1.746          & 3.233          & 4.026          & 6.080          & 2.981          & 4.913          \\ 
S2TNet \cite{chen2021s2tnet} & \textbf{1.1679} & \textbf{1.9874} & \textbf{0.6834}	& \textbf{1.700} & 2.1798 & 3.5783 & 1.3048 & 3.2151\\ \hline
STF    & 1.384 & 2.403 & 0.799 & 2.001 & \textbf{1.707} & \textbf{2.972} & \textbf{1.012} & \textbf{2.392} \\ \hline
\end{tabular}
}

\end{table}

     Root Mean Square Error (RMSE): measures the distance between the predictor and the ground truth at time $t$ \cite{song2020pip}.

    \begin{equation}
        RMSE^t=\sqrt{\frac{\sum_{n\in \mathbb{N}}\left(\hat{y}_{n}^t-y_{n}^t\right)^2}{N}}
        \label{eq:rmse}
    \end{equation}
    The Apollo Scape Trajectory dataset provides a weighted sum formula for RMSE, given by Equation (\ref{eq:weighted rmse}).
    \begin{equation}
        RMSE_w = D_v * RMSE_v + D_p * RMSE_p + D_b * RMSE_b
        \label{eq:weighted rmse}
    \end{equation}  
where $D_v = 0.20$,$D_p = 0.58$, and $D_b = 0.22$. ApolloScape Trajectory Dataset provides these values and points out that they are computed based on the velocity information available in the dataset. $RMSE_v$ represents the RMSE value for vehicles, $RMSE_p$ represents the RMSE value for pedestrians, and $RMSE_b$ represents the RMSE value for bikes.

     Average Displacement Error (ADE): calculates the distance between the predicted location of $N$ agents and ground truth in  $T_{pred}$ time steps, which measures the average prediction performance along the trajectory \cite{deoConvolutionalSocialPooling2018a-cslstm}
    \begin{equation}
        ADE=\frac{\sum _{t \in \mathbb{T}}\sum_{n\in \mathbb{N}}\|\hat{y}_{n}^t-y_{n}^t\|_2}{N\times T_{pred}}
    \end{equation}  
    where $\hat{y}_{n}^t$ is the prediction location of $nth$ vehicle at time step $t$, $y_{n}^t$ is the true location of $nth$ vehicle at time step $t$. $\mathbb{T}=\{T_{his}+1, T_{his}+2, ..., T_{his}+T_{pred}\}$ and $\mathbb{N}=\{1,2,...,N\}$. $ADE_v$,$ADE_p$ and $ADE_b$ are computed separately.
    
     Final Displacement Error (FDE): only computes the distance between the predicted final location  and the true final location at the end of final prediction time step $T_{pred}$ \cite{deoConvolutionalSocialPooling2018a-cslstm}
    \begin{equation}
        FDE=\frac{\sum_{n\in \mathbb{N}}\|\hat{y}_{n}^{T_{pred}}-y_{n}^{T_{pred}}\|_2}{N}
    \end{equation}
    where $\hat{y}_{n}^{T_{pred}}$ is the prediction location of $nth$ vehicle at the last prediction time step $T_{pred}$, $y_{n}^{T_{pred}}$ is the true location of $nth$ traffic participant at the last prediction time step $T_{pred}$. $FDE_v$, $FDE_p$ and $FDE_b$ are computed separately.
The WSADE and WSFDE can be computed through Equation (\ref{eq:wsade}) and Equation (\ref{eq:wsfde}). 
    \begin{equation}
        WSADE = D_c * ADE_v + D_p * ADE_p + D_b * ADE_b
        \label{eq:wsade}
    \end{equation}
    \begin{equation}
        WSFDE = D_c * FDE_v + D_p * FDE_p + D_b * FDE_b
        \label{eq:wsfde}
    \end{equation}
where the $D_v = 0.20$,$D_p = 0.58$, and $D_b = 0.22$ which is the same as them in Equation (\ref{eq:weighted rmse}) provided by the ApolloScape Trajectory Dataset.
\subsection{Experiments on the ApolloScape Trajectory Datasets}

\begin{table}[htbp]
\centering
\caption{RMSE for trajectory prediction on Apollo Scape Trajectory}
\label{tab:result}
\resizebox{\columnwidth}{!}
{
\begin{tabular}{|c|c|cccccc|}
\hline
 & {Object type} & \multicolumn{6}{c|}{RMSE} \\
\cline{3-8}
& & 0.5s & 1s & 1.5s & 2s & 2.5s & 3s \\
\hline
& vehicle & \textbf{1.744} & 2.577 & 3.513 & 4.440 & 5.335 & 6.080 \\
& pedestrian & 0.515 & 1.002 & 1.522 & 2.022 & 2.436 & 2.981\\
{GRIP++} & bike & \textbf{1.101} & 2.145 & 2.924 & 3.779 & 4.535 & 4.913  \\
& weighted RMSE & \textbf{0.890} & 1.569 & 2.228 & 2.892 & 3.478 & 4.026 \\
& all objects & \textbf{1.384} & 2.218 & 3.033 & 3.919 & 4.722 & 5.425 \\
\hline
& vehicle & 1.972 & \textbf{2.0} & \textbf{2.311} & \textbf{2.517} & \textbf{2.649} & \textbf{2.972} \\
& pedestrian & \textbf{0.50 }& \textbf{0.653} & \textbf{0.773} & \textbf{0.887} & \textbf{0.971} & \textbf{1.012} \\
{STF} & bike & 1.366 & \textbf{1.833} & \textbf{1.925} & \textbf{2.207} & \textbf{2.282} & \textbf{2.392} \\
& weighted RMSE & 0.985 & \textbf{1.182} & \textbf{1.334} & \textbf{1.503} & \textbf{1.595} & \textbf{1.707} \\
& all objects & 1.586 & \textbf{1.792} & \textbf{2.029} & \textbf{2.265} & \textbf{2.363} & \textbf{2.612} \\
\hline
\end{tabular}
}
\end{table}

Table \ref{tab:ade-fde} shows the competition results of TafficPredict, GRIP++, S2TNet, and STF. The metrics reported include $WSADE$, $ADE_v$, $ADE_p$, $ADE_b$, $WSFDE$, $FDE_v$, $FDE_p$, and $FDE_b$. Among the listed methods, it is evident that the STF model demonstrates competitive performance across several metrics.
For the $WSADE$ metric, STF achieves a value of 1.384, which is lower than TafficPredict (8.5881) and GRIP++ (2.514), indicating improved accuracy. Similarly, for the $ADE_v$, $ADE_p$, and $ADE_b$ metrics, the STF model outperforms TafficPredict and GRIP++, showing lower values and better accuracy in trajectory prediction. But the higher values than the result of S2TNet, which means the S2TNet has better performance than STF in short-time prediction.
Moving on to the $WSFDE$, $FDE_v$, $FDE_p$, and $FDE_b$ metrics, the STF model excels, demonstrating the lowest values among the listed methods. Specifically, STF achieves an impressive $WSFDE$ of 1.707, an $FDE_v$ of 2.972, an $FDE_p$ of 1.012, and an $FDE_b$ of 2.392. These results indicate that the STF model can predict long-term trajectories with greater precision and accuracy compared to TafficPredict, GRIP++, and S2TNet.

In summary, the STF model stands out as the best-performing method in trajectory prediction based on the competition results. It surpasses TafficPredict, GRIP++, and S2TNet in terms of $FDE$ metrics, demonstrating superior accuracy and precision in trajectory prediction on long-time trajectory prediction.

Due to the lack of $RMSE$ values at every frame in the TafficPredic and S2TNet models, we evaluate our model by comparing it with GRIP++ based on the $RMSE$ at each prediction frame. Furthermore, our model incorporates GRIP++ as its backbone. The outcomes derived from GRIP++ can be regarded as an ablation experiment.
The experiment results, as shown in Table \ref{tab:result}, compare the $RMSE$ for trajectory prediction on the Apollo Scape Trajectory dataset between the GRIP++ model and the STF. The RMSE values are reported for different prediction time intervals, ranging from 0.5s to 3s.

\begin{figure}
    \centering
    \includegraphics[width = 0.8\linewidth]{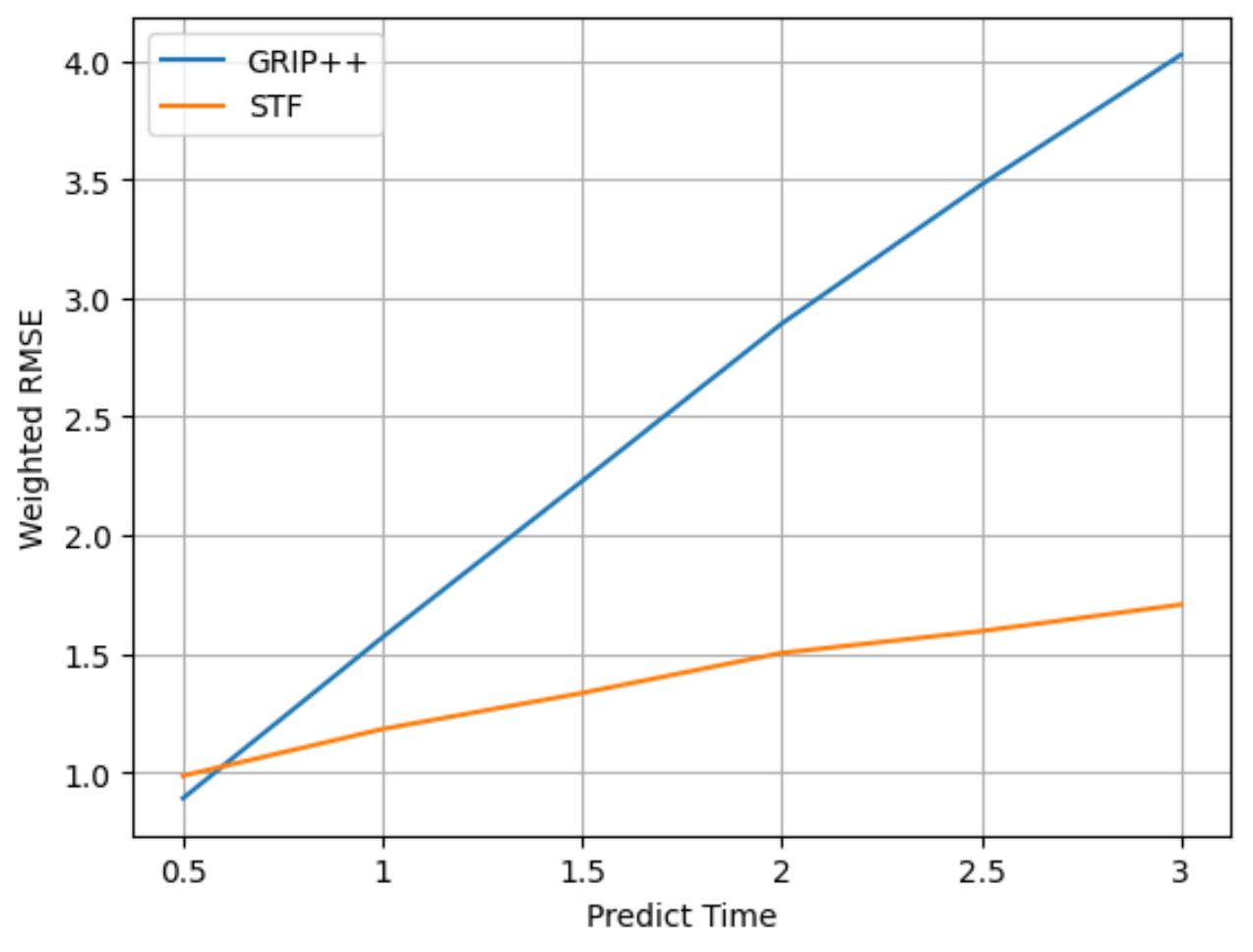}
    \caption{Weighted RMSE in prediction time}
    \label{fig:rmse-plot}
\end{figure}

Overall, the STF outperforms the GRIP++ model in terms of trajectory prediction accuracy. Specifically, the STF achieves lower $RMSE$ values for most object types and prediction time intervals compared to GRIP++.
For the car object type, the STF obtains slightly higher $RMSE$ values at shorter prediction time intervals (0.5s) but achieves better performance at longer prediction time intervals (1s, 1.5s, 2s, 2.5s, and 3s). For pedestrians, the STF consistently exhibits lower $RMSE$ values across all prediction time intervals, especially for long-term prediction. For the bike object type, the STF achieves better results at longer prediction time intervals (1s, 1.5s, 2s, 2.5s, and 3s), while the $RMSE$ values are comparable or slightly higher at shorter intervals (0.5s). For the weighted sum of these three kinds of objects computed according to Equation (\ref{eq:weighted rmse}), the STF shows much better on the long-term (1s, 1.5s, 2s, 2.5s, and 3s) trajectory prediction.
The STF model demonstrates superior performance compared to the GRIP++ model in terms of all object types. This includes vehicles, pedestrians, bikes, and other objects. The STF consistently achieves lower $RMSE$ values across long prediction time intervals (1s, 1.5s, 2s, 2.5s, and 3s).
Fig. \ref{fig:rmse-plot} displays the plot of the Weighted $RMSE$ for GRIP++ and STF at various predicted times. This figure highlights the superior performance of STF in accurately predicting future trajectories over an extended period.

\section{Conclusion}
This paper presents the STF model, designed to capture spatial and temporal information from historical trajectory data simultaneously. The model addresses the information missing problem in previous works while mitigating the issue of error accumulation. Experiments are conducted on the ApolloScape Trajectory Datasets. The results demonstrate superior performance compared to previous models. Our model exhibits a significant advantage, particularly in long-term trajectory prediction performance. Accurate long-term predictions ensure that autonomous vehicles and robotics can proactively avoid potential collisions, providing them with more time to respond, ultimately resulting in enhanced safety.

\bibliographystyle{IEEEtran}
\bibliography{1reference}


\end{document}